\PassOptionsToPackage{dvipsnames}{xcolor}
\documentclass{article} 
\usepackage[preprint]{colm2025_conference}

\usepackage{microtype}
\usepackage{hyperref}
\usepackage{url}
\usepackage{booktabs}
\usepackage{colortbl}
\usepackage{array}
\usepackage{lineno}
\usepackage{csquotes}
\usepackage{amsmath,amssymb,amsthm} 
\usepackage{times}
\usepackage{latexsym}
\usepackage{multirow}
\usepackage{amsfonts}
\usepackage{graphicx} 
\usepackage{pifont}
\usepackage{enumitem}
\usepackage{tikz}
\usepackage[skins]{tcolorbox}
\usepackage{wrapfig}
\usepackage{subfigure}
\usepackage{titletoc}

\usepackage[normalem]{ulem}
\newtheorem{definition}{Definition}
\newcommand{\cellbg}[2][green!20]{%
  \begin{tikzpicture}[baseline=(text.base)]
    \node[fill=#1, inner sep=2pt, outer sep=0.6pt, anchor=base] (text) {#2};
  \end{tikzpicture}%
}

\usepackage{subcaption}
\definecolor{darkblue}{rgb}{0, 0, 0.5}
\definecolor{deepred}{rgb}{0.9, 0.2, 0.2}
\definecolor{medred}{rgb}{1, 0.6, 0.6}
\definecolor{lightred}{rgb}{1, 0.8, 0.8}
\definecolor{deepgreen}{rgb}{0.2, 0.8, 0.2}
\definecolor{medgreen}{rgb}{0.6, 0.9, 0.6}
\definecolor{lightgreen}{rgb}{0.8, 1, 0.8}

\hypersetup{colorlinks=true, citecolor=darkblue, linkcolor=darkblue, urlcolor=darkblue}

\title{Missing Premise exacerbates Overthinking:\\Are Reasoning Models losing Critical Thinking Skill?}

\author{Chenrui Fan\textsuperscript{1*}, 
Ming Li\textsuperscript{1*}, 
Lichao Sun\textsuperscript{2}, 
Tianyi Zhou\textsuperscript{1} \\
  \textsuperscript{1}University of Maryland; 
  \textsuperscript{2}Lehigh University\\
  \texttt{\{cfan42, minglii, tianyi\}@umd.edu} \\
  Project: \url{https://github.com/tianyi-lab/MiP-Overthinking}
  }

\newcommand{\header}[1]{\noindent\textbf{#1}.}

\begin{document}

\ifcolmsubmission
\linenumbers
\fi

\maketitle

\renewcommand{\thefootnote}{}
\footnotetext{*Equal Contribution.}
\renewcommand{\thefootnote}{\arabic{footnote}}

\begin{displayquote}
"The Answer to the Great Question... Of Life, the Universe and Everything... is... \textbf{\textit{Forty-two}}," said Deep Thought, with infinite majesty and calm.
\end{displayquote}
    
\begin{flushright}
\begin{minipage}{0.5\textwidth}
--- \textit{The Hitchhiker's Guide to the Galaxy}
\end{minipage}
\end{flushright}

\begin{abstract}
We find that the response length of reasoning LLMs, whether trained by reinforcement learning or supervised learning, drastically increases for ill-posed questions with missing premises (MiP), ending up with redundant and ineffective thinking. 
This newly introduced scenario exacerbates the general overthinking issue to a large extent, which we name as the MiP-Overthinking. 
Such failures are against the ``test-time scaling law'' but have been widely observed on multiple datasets we curated with MiP, indicating the harm of cheap overthinking and a lack of critical thinking. 
Surprisingly, LLMs not specifically trained for reasoning exhibit much better performance on the MiP scenario, producing much shorter responses that quickly identify ill-posed queries. This implies a critical flaw of the current training recipe for reasoning LLMs, which does not encourage efficient thinking adequately, leading to the abuse of thinking patterns. 
To further investigate the reasons behind such failures, we conduct fine-grained analyses of the reasoning length, overthinking patterns, and location of critical thinking on different types of LLMs. 
Moreover, our extended ablation study reveals that the overthinking is contagious through the distillation of reasoning models' responses.
These results improve the understanding of overthinking and shed novel insights into mitigating the problem. 
\looseness-1
\end{abstract}

\section{Introduction}

    




Reasoning abilities in large language models (LLMs) have become a cornerstone of advanced AI applications~\citep{huang2023reasoninglargelanguagemodels, li2024happened, ahn-etal-2024-large, wang2025multimodalchainofthoughtreasoningcomprehensive}, powering breakthroughs in mathematical reasoning~\citep{xiong2025selfrewardingcorrectionmathematicalreasoning, xia2025evaluatingmathematicalreasoningaccuracy}, code generation~\citep{liu2024codemindframeworkchallengelarge}, and commonsense question answering~\citep{wang2023geminireasoningunveilingcommonsense}. 
These gains often stem from the scaling law of model/dataset sizes~\citep{kaplan2020scalinglawsneurallanguage} in both pre-training ~\citep{shao2024deepseekmathpushinglimitsmathematical} and post-training, which unlock emergent capabilities such as step-by-step reasoning and reflection skills witnessed on  
OpenAI’s GPT-o1~\citep{openai2024o1} and the open-source DeepSeek-R1~\citep{deepseekai2025deepseekr1incentivizingreasoningcapability}. 
By leveraging supervised fine-tuning (SFT) on expert responses~\citep{ye2025limoreasoning, muennighoff2025s1simpletesttimescaling} and/or reinforcement learning (RL)~\citep{deepseekai2025deepseekr1incentivizingreasoningcapability}, these models are tailored to produce detailed multi-step reasoning paths, whose length increase usually associated with improved performance on complex tasks such as math reasoning and programming. 

Despite the fascinating reasoning capabilities exhibited on recent models, there is growing concern about the efficiency and quality of the long reasoning process~\citep{sui2025stopoverthinkingsurveyefficient}. 
\cite{chen2025think23overthinkingo1like} first raises the ``overthinking'' problem in reasoning LLMs, which is reflected by the excessively long reasoning paths generated for extremely simple queries. 
For example, even for questions like ``\textit{What is the answer of 2 plus 3?}'', existing reasoning models might generate hundreds of response tokens. 

\begin{figure*}[t]
    \centering
    \includegraphics[width=\linewidth]{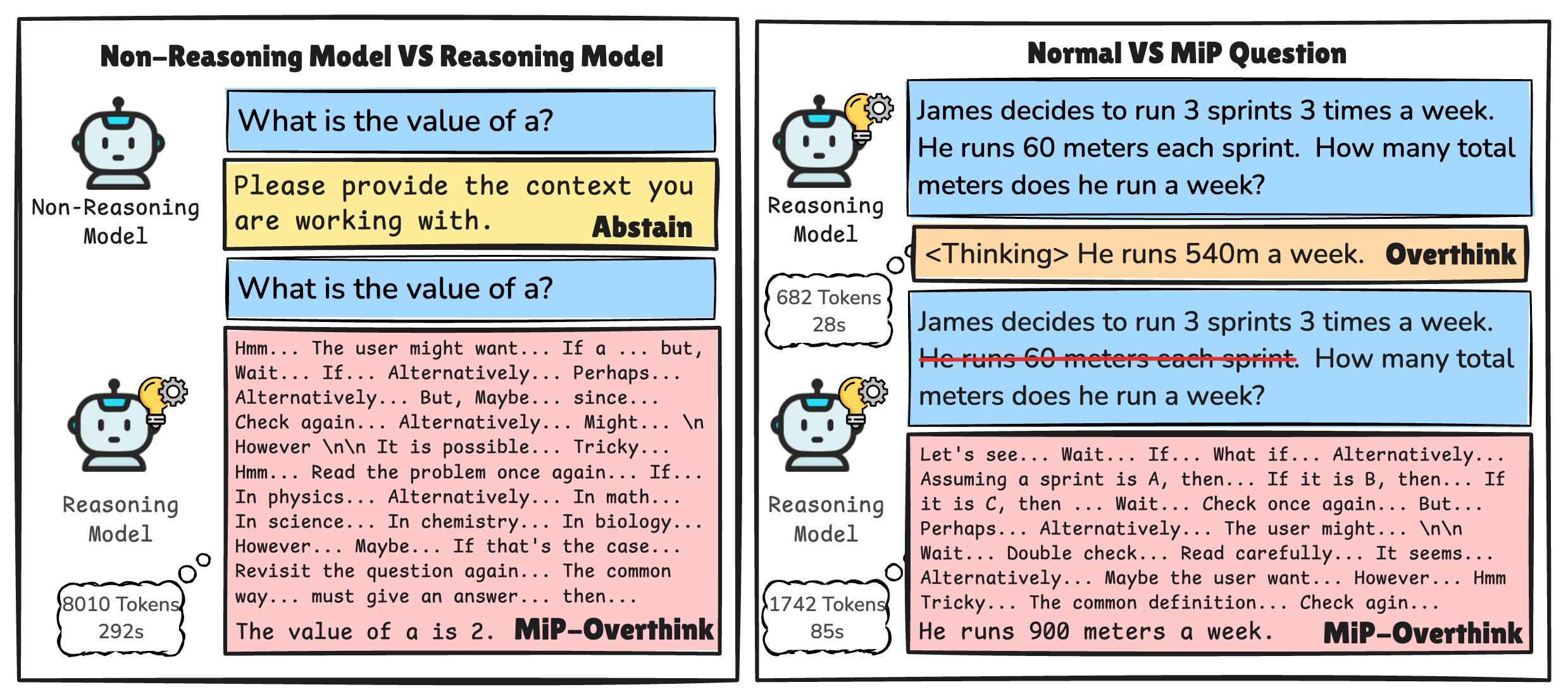}
    \vspace{-8mm}
    \caption{Illustration of MiP-Overthinking. When queried by questions with missing premises, the response length of reasoning models increases excessively, and they cannot abstain from answering with MiP identified. The left shows a query with an undefined variable, while the right compares a well-defined GSM8K question with its MiP variant (with a critical numerical condition removed). Reasoning models' responses to MiP questions are much longer than those for well-defined questions and those generated by non-reasoning models. The left corner of each response report the response length and thinking time by DeepSeek-R1.}
    \label{fig:illustration}
    \vspace{-6mm}
\end{figure*}

In particular, the ill-posed queries are unsolvable due to the lack of a necessary premise or condition. We call the reasoning failure for the ill-posed queries \textbf{Overthinking under Missing Premise (MiP-Overthinking)}.
For example, the simplest MiP question is \textit{What is the value of $a$?}
\footnote{In \textit{The Hitchhiker's Guide to the Galaxy}, the supercomputer Deep Thought spends hundreds of years to answer the \textit{the Ultimate Question of Life, the Universe, and Everything} as \textbf{42}, and we observe that DeepSeek-R1 spends thousands of tokens to answer \textit{What is the value of $a$} as \textbf{2}, which we find them interestingly alike. }
, as shown on the left part of Figure \ref{fig:illustration}. Without providing any other information regarding $a$, it is evidently unsolvable. 
However, DeepSeek-R1 generates thousands of tokens and spends several minutes thinking about this question before outputting the final meaningless answer. 
In this paper, we find that a trivial type of ill-posed queries will significantly exacerbate the overthinking of reasoning models, resulting in excessively redundant and meaningless thinking. In contrast, humans and even non-reasoning models are often immune to such scenarios and quickly end up by questioning the validity of the given query, indicating the critical thinking capability. 
This exposes a risk of the abuse of thinking patterns and a lack of critical thinking on the models trained for deep thinking. 
Ideally, a model with critical thinking skills is expected to identify the missing premise and quickly respond  
by either requesting clarification or gracefully indicating that it cannot proceed~\citep{cole2023selectivelyansweringambiguousquestions, amayuelas2024knowledgeknowledgeexploringknownunknowns}.

MiP-Overthinking differs from the widely discussed overthinking issue~\citep{cuadron2025dangeroverthinkingexaminingreasoningaction}, in which the query is usually well-defined, but a model applies much more reasoning than necessary for little benefit. 
MiP-Overthinking, by contrast, happens when the question itself is ill-posed and lacks sufficient information to be solved. 
For example, the right of Figure~\ref{fig:illustration} presents a well-defined question from GSM8K and a MiP variant, where the latter triggers a drastic increase of the generated tokens on recent reasoning models compared with the general overthinking. 
\textbf{Overthinking can be presented by the length difference between models addressing the same well-defined questions, while MiP-Overthinking can be presented by the additional tokens generated due to MiP.}
MiP-Overthinking further reveals the lack of critical thinking that questions the validity of ill-posed questions and quickly identifies MiP, thus abstaining from answering the questions.
Moreover, we observe that reasoning models' ineffective and redundant thinking often cannot stop even after successful notice of MiP, violating the expectation of test-time scaling law. 
Hence, MiP-Overthinking indicates potential drawbacks of current training recipes of reasoning models. 

To systematically investigate this issue, we construct a suite of MiP questions designed to trigger the overthinking failures in a controlled way. These include synthetic questions generated by Rule-based Formula (queries where a formula reference is empty or nonsensical) and careful modifications of established datasets across diverse levels of difficulties, including SVAMP, GSM8K, and MATH500.
On the modified datasets of MiP questions, we empirically evaluate a wide range of state-of-the-art LLMs, from reasoning models to non-reasoning models and from open-sourced models to proprietary models, to ensure the generalizability of our findings. 
Our analysis is mainly based on three evaluation metrics, the length of generated responses, the accuracy on well-defined questions, and the abstain rate on ill-posed questions with MiP. 



\textbf{Main Contributions:}
We present the first in-depth study of \textbf{\textit{Overthinking under Missing Premise (MiP-Overthinking)}}, 
which reveals a critical shortcoming in existing reasoning models: Although they appear to follow coherent 
reasoning patterns, \textbf{they lack genuine critical thinking capabilities}. To systematically analyze this issue, 
we curate four MiP datasets covering various difficulty levels and three ill-posed question generation strategies, i.e., \textit{Rule-Based Generation, Body-Question Swapping, and Essential-Premise Removal.} We then evaluate a wide range of large language 
models including reasoning-based and non-reasoning ones. Our empirical results illuminate the differences in how 
models handle well-defined vs. MiP questions, ultimately offering insights into the limitations of existing reasoning models.

\textbf{Our key findings:}
\begin{enumerate}[label=\textbf{\arabic*.}, leftmargin=4mm]
    \item Missing premise in questions induces reasoning models to generate significantly longer ($2\times$ to $4\times$ more tokens) responses than general overthinking on well-defined questions. The increased tokens fail to help identify MiP in the ill-posed questions, surprisingly \textbf{contradicting the widely-discussed test-time scaling law}.
    \item In contrast, given MiP questions, \textbf{non-reasoning models generate consistently shorter responses and quickly identify MiP}, demonstrating greater robustness to the absence of critical information. 
    \item Reasoning models respond differently to well-defined vs. MiP questions: they mostly follow stable chain-of-thoughts for the former, but are often \textbf{trapped in a self-doubt loop, repeatedly revisiting the question, and guessing the user intentions} under MiP, resulting in an explosion of tokens. \looseness-1
    \item Reasoning models often can \textbf{notice the existence of MiP or identify it at an early stage}, but they \textbf{hesitate to commit to this judgment} and keep outputting ineffective thinking. 

\end{enumerate}
\section{Missing Premise Definition and Construction}

\subsection{Definition of Missing Premise}
Prior to introducing the construction our dataset and analyzing the behavior of reasoning models on problems with missing premises, we formally define the Missing Premise (MiP) problem to establish a rigorous foundation for our subsequent analysis.

\begin{tcolorbox}
\begin{definition}[Missing Premise Problem]\label{def:mip}
    Let $\mathcal{Q}$ be a question, and let $P = \{P_1, \dots, P_n\}$ be a set of premises. Define the function mapping premises and a question to the set of logically valid answers as:
    \begin{equation}
    \mathcal{F}(P, \mathcal{Q}) = \{A \mid P \vdash A, A \text{ is an answer resolving } \mathcal{Q}\}
    \end{equation}
    where $\vdash$ denotes logical entailment. Consider a proper subset $P' = P \setminus \{P_i\}$ for some $P_i \in P$. The tuple $(P', \mathcal{Q})$ forms a \textbf{missing premise problem} if and only if:
    \begin{equation}
    |\mathcal{F}(P, \mathcal{Q})| = 1 \quad \text{and} \quad |\mathcal{F}(P', \mathcal{Q})| \neq 1
    \end{equation}
    This indicates that the removed premise $P_i$ is essential for uniquely determining the logically valid answer to the question $\mathcal{Q}$.
\end{definition}
\end{tcolorbox}

According to Definition~\ref{def:mip}, an ideal reasoning system should efficiently identify the absence of a critical premise and terminate its inference process upon recognizing that the available information is insufficient to derive a unique solution to the given problem. However, our empirical analysis in Section~\ref{sec:main} demonstrates that state-of-the-art reasoning models consistently fail to exhibit this capability. Instead, these models engage in extensive, redundant reasoning chains that consume significant computational resources without ultimately identifying the missing premise. 

\subsection{Overview of Data Construction}

\renewcommand{\arraystretch}{0.8}

\begin{table}[t]
    \centering
    \resizebox{1.0\textwidth}{!}{%
    \begin{tabular}{
        m{0.14\textwidth}
        m{0.52\textwidth}
        >{\centering\arraybackslash}m{0.05\textwidth}
        >{\centering\arraybackslash}m{0.05\textwidth}
        >{\centering\arraybackslash}m{0.05\textwidth}
        >{\centering\arraybackslash}m{0.15\textwidth}
    }
    \toprule
    Dataset & Example & Diff & Count & Pair & Method\\
    \midrule
    MiP-Formula 
      & What is the value of $\ln(a + b)$? 
      & {\Large $\star$} 
      & 50 
      & \ding{55} & Rule-Based Generation\\ \midrule
    MiP-SVAMP 
      & Paco had 26 salty cookies and 17 sweet cookies.
        He ate 14 sweet cookies and 9 salty cookies. \sout{How many salty cookies did Paco have left?} How many pencils does she have?
      & {\Large $\star$} 
      & 300 
      & \ding{55} & Body-Question Swapping \\ \midrule
    MiP-GSM8K 
      & James decides to run 3 sprints 3 times a week. \sout{He runs 60 meters each sprint.}
        How many total meters does he run a week?
      & {\Large $\star\star$} 
      & 582 
      & \ding{51} & Essential-Premise Removal\\ \midrule
    MiP-MATH 
      & There are 360 people in my school. 15 take calculus, physics, and chemistry, 
        and 15 don't take any of them. 180 take calculus. Twice as many students take chemistry as take physics. 75 take both calculus and chemistry, and 75 take both physics and chemistry.  \sout{Only 30 take both physics and calculus.} How many students take physics?
      & {\Large $\star\star\star$} 
      & 58 
      & \ding{51} & Essential-Premise Removal\\
    \bottomrule
    \end{tabular}
    }
    \vspace{-2mm}
    \caption{Statistics and examples of our curated MiP datasets. For GSM8K and MATH, a premise is removed from the original questions (crossed out) to create MiP questions. 
    \textit{Diff} represents the (estimated) difficulty for models to identify MiP. \textit{Count} denotes the number of questions in the subset. \textit{Pair} indicates whether each MiP question is associated with a well-defined original question. \textit{Method} indicates the method used to generate the MiP question.}
    \vspace{-2mm}
    \label{tab:data_samples}
\end{table}

To systematically investigate this MiP-Overthinking issue, we construct a suite of MiP questions in a controllable manner. Our MiP questions are sourced from $3$ math datasets across different difficulties. In addition, we also construct a synthetic dataset consisting of formulas with unassigned variables. Our ill-posed question generation employs three distinct methods covering three difficulty levels and three strategies to create MiP questions:
\begin{itemize}[leftmargin=2em]
    \item \textbf{Rule-Based Generation}: This approach generates MiP questions through a principled formula construction process, where unassigned variables serve as the missing premises. 
    
    \item \textbf{Body-Question Swapping}: We introduce logical inconsistencies by deliberately mismatching problem bodies with their corresponding questions from the original dataset. This creates scenarios where the premises and queries are fundamentally incompatible.
    
    \item \textbf{Essential-Premise Removal}: Through careful analysis of existing well-formed questions, we identify and remove critical premises that are necessary for logical resolution. This transformation preserves the question's structure while rendering it unsolvable.
\end{itemize}

The following sections provide a detailed overview of our data construction process for each dataset category. For comprehensive implementation details and additional methodological considerations, we refer readers to Appendix~\ref{sec:data_detail}.

\header{MiP-Formula}
We construct a dataset of $50$ synthetic unsolvable formulas in a rule-based manner. The formulas are generated recursively through combinations of variables and operators, with a maximum recursion depth of three. 
While these formulas may appear complex at a glance, their unsolvability should be immediately apparent due to the presence of undefined variables.

\header{MiP-SVAMP}
We utilize SVAMP~\citep{patel-etal-2021-nlp}, a benchmark dataset with elementary-school-level math problems, where each instance consists of a problem body and an associated question. 
We generate MiP question by randomly permuting the problem bodies and associated questions and then manually inspect them to avoid inadvertent cases. 
The resulting problems contain clear logical inconsistencies between their body and question components, which is easy for a human to identify.

\header{MiP-GSM8K}
We further utilize GSM8K~\citep{cobbe2021gsm8k}, a more complex mathematics dataset than SVAMP. 
The questions in GSM8K typically contain multiple numerical conditions and require certain reasoning capabilities to arrive at solutions. 
We first identify the questions containing two or three numerical conditions and then randomly eliminate one numerical condition per question before conducting human verification to filter out those questions that are still solvable in some way. 
Compared with previous MiP questions, questions from this source require the basic logical analysis of models to identify that the question is unsolvable.  

\header{MiP-MATH}
For MATH 500 dataset~\citep{hendrycksmath2021}, which contains challenging mathematical questions at the competition level, it is difficult to build a rule-based filtering mechanism. Thus, we manually select $58$ questions that are feasible for constructing the MiP questions and remove one necessary premise from the question. 
Due to the sophisticated nature of this data source, identifying the insufficiency of these instances requires substantial mathematical reasoning capabilities, testing models' ability to recognize unsolvability in complex mathematical contexts.



\section{Overthinking under Missing Premise}



\begin{figure*}[t]
    \centering
    \includegraphics[width=\linewidth]{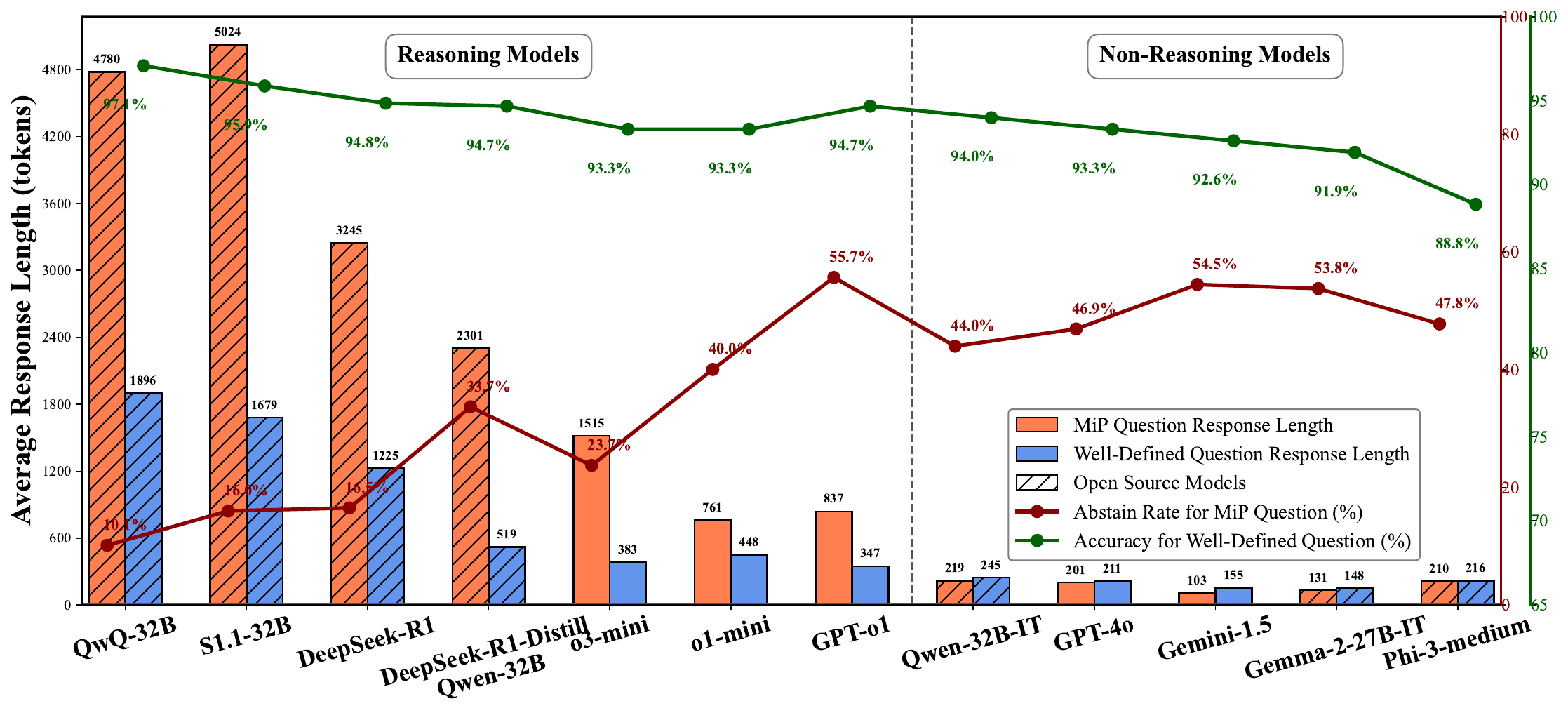}
    \vspace{-8mm}
    \caption{Response lengths, accuracy on well-defined questions, and abstain rate of reasoning/non-reasoning models on MiP questions from our MiP-GSM8K dataset. 
    (1) Existing reasoning models generate significantly longer responses for MiP questions than well-defined questions, while non-reasoning models generate responses of similar lengths for both types of questions, indicating \textbf{MiP-Overthinking} for reasoning models. 
    (2) For both questions, reasoning models generate longer responses than non-reasoning models, indicating \textbf{General Overthinking}. 
    (3) Although the longer responses by reasoning models slightly improve the accuracy for well-defined questions, it does not enhance the abstain rate for MiP questions, indicating a \textbf{contradiction on the test-time scaling law}. 
    }
    \label{fig:mian_fig}
    \vspace{-4mm}
\end{figure*}

\subsection{Evaluation Metrics}
To systematically evaluate model responses under MiP, we conduct experiments with a diverse set of reasoning and non-reasoning models. For each model, we analyze calculate the following metrics for the responses across different datasets:
\vspace{-8pt}
\begin{itemize}[leftmargin=4mm]
    \item \textbf{Response Length}: The average number of tokens in the response, incorporating both reasoning steps and final answer components.
    \item \textbf{Abstain Rate for MiP Question}: The proportion of answers where the model explicitly identifies the missing premise and either declines to provide an answer or requests additional information necessary for solving the problem.
    \item \textbf{Accuracy for Well-defined Question}: The proportion of answers where the model produces a definitive response that aligns with the reference answer.
\end{itemize}
\vspace{-8pt}
For datasets without reference answers (MiP-Formula and MiP-SVAMP), we only calculate the abstain rate for the questions. Response evaluation is performed using GPT-4o as an automated evaluator. Detailed experimental procedures and evaluation protocols are provided in Appendix~\ref{sec:experiment}.

\subsection{Main Results}
\label{sec:main}

Figure \ref{fig:mian_fig} compares average response length, accuracy on well-defined questions, and the abstain rate on MiP questions across a range of state-of-the-art LLMs, revealing several significant patterns in model behavior.

Firstly, existing reasoning models (left side of the figure) display an explosive increase in response length when facing the MiP questions, often producing $2$ – $4\times$ more tokens than general overthinking on well-defined questions. 
For example, QwQ-32B~\citep{qwq32b} and DeepSeek-R1~\citep{deepseekai2025deepseekr1incentivizingreasoningcapability} exhibit a substantial increase from already long reasoning paths on well-defined questions (approximately $1,000$ tokens for simple GSM8K questions) to highly lengthy outputs (more than $3,000$ tokens) under missing premise conditions. 
On the contrary, no similar issues exist for non-reasoning models (right side of the figure), which generate similar token counts for both types of well-defined and MiP questions.  
This phenomenon directly illustrates the \textbf{NiP-Overthinking} phenomenon as introduced in the paper. 

Secondly, comparing the token lengths on well-defined questions between the reasoning and non-reasoning models, reasoning models tend to produce longer responses, even for simple questions, than non-reasoning models, underscoring the inefficient and verbose responses of existing reasoning models. 
For example, for the non-reasoning models, it only takes approximately $200$ tokens for them to generate the responses for well-defined questions, while taking $1,000$ tokens for DeepSeek-R1 and $1,800$ tokens for QWQ-32B to answer the exactly same questions.  
However, the explosive increase in extra tokens does not lead to corresponding large accuracy improvements, shown in the green line, highlighting the issue of the \textbf{General Overthinking}. 


Finally, the abstain rates (red line) on MiP questions reveal that although some reasoning models (e.g., GPT-o1) have promising capabilities in abstaining from the MiP questions, most of the other reasoning models are not able to abstain from the given MiP questions correctly despite the dramatically long reasoning paths. 
This phenomenon indicates that although most existing reasoning models have thinking and reasoning capabilities to some extent, they \textbf{lack the critical thinking capabilities} to ``reject'' ill-posed questions. 
By contrast, non-reasoning models, though they are not explicitly trained for reasoning, tend to strike a better balance, generating shorter answers that are more likely to acknowledge MiP when the question is ill-posed. 
This phenomenon reveals a surprising \textbf{contradiction on test-time scaling law}. 

\begin{table*}[t]
\centering
\renewcommand\arraystretch{0.6}
\resizebox{\textwidth}{!}{%
\begin{tabular}{l|c|cc|cc|c|cc|cc}
\toprule
\multirow{2}{*}{\textbf{Model}} & \multirow{2}{*}{\textbf{Type}} & \multicolumn{2}{c|}{\textbf{MiP-Formula}} & \multicolumn{2}{c|}{\textbf{MiP-SWAMP}} & \multirow{2}{*}{\textbf{Type}} & \multicolumn{2}{c|}{\textbf{MiP-GSM8K}} & \multicolumn{2}{c}{\textbf{MiP-MATH}} \\
& & Length$\downarrow$& Abstain$\uparrow$ & Length$\downarrow$& Abstain$\uparrow$ & & Length$\downarrow$& Abstain$\uparrow$ & Length$\downarrow$& Abstain$\uparrow$ \\
\midrule
\multicolumn{11}{c}{\textbf{Non-Reasoning Models}} \\ \midrule
\multirow{2}{*}[-0.5\baselineskip]{\textbf{Qwen2.5-32B-Instruct}} & \multirow{2}{*}[-0.5\baselineskip]{MiP} & \multirow{2}{*}[-0.5\baselineskip]{\cellbg[lightgreen]{285}} & \multirow{2}{*}[-0.5\baselineskip]{44.0} & \multirow{2}{*}[-0.5\baselineskip]{128} & \multirow{2}{*}[-0.5\baselineskip]{98.3} & MiP & 219 & 44.0 & 525 & 15.4 \\
& & & & & &  Well-defined & 246 & 0.5 & 1114 & 1.9 \\
\arrayrulecolor{gray}\midrule
\multirow{2}{*}[-0.5\baselineskip]{\textbf{GPT-4o}} & \multirow{2}{*}[-0.5\baselineskip]{MiP} & \multirow{2}{*}[-0.5\baselineskip]{\cellbg[lightgreen]{338}} & \multirow{2}{*}[-0.5\baselineskip]{70.0} & \multirow{2}{*}[-0.5\baselineskip]{\cellbg[lightgreen]{122}} & \multirow{2}{*}[-0.5\baselineskip]{96.3} & MiP & \cellbg[lightgreen]{202} & 46.9 & 487 & 15.4 \\
& & & & & &  Well-defined & 212 & 0.5 & 472 & 1.9 \\
\midrule
\multirow{2}{*}[-0.5\baselineskip]{\textbf{Gemini 1.5}} & \multirow{2}{*}[-0.5\baselineskip]{MiP} & \multirow{2}{*}[-0.5\baselineskip]{453} & \multirow{2}{*}[-0.5\baselineskip]{{20.0}} & \multirow{2}{*}[-0.5\baselineskip]{\cellbg[deepgreen]{52}} & \multirow{2}{*}[-0.5\baselineskip]{\cellbg[deepgreen]{99.0}} & MiP & \cellbg[deepgreen]{103} & \cellbg[deepgreen]{54.5} & 568 & \cellbg[lightred]{5.8} \\
& & & & & &  Well-defined & 156 & 0.5 & 502 & 0.0 \\
\midrule
\multirow{2}{*}[-0.5\baselineskip]{\textbf{Gemma-2-27B-IT}} & \multirow{2}{*}[-0.5\baselineskip]{MiP} & \multirow{2}{*}[-0.5\baselineskip]{\cellbg[deepgreen]{204}} & \multirow{2}{*}[-0.5\baselineskip]{\cellbg[deepgreen]{85.7}} & \multirow{2}{*}[-0.5\baselineskip]{\cellbg[medgreen]{89}} & \multirow{2}{*}[-0.5\baselineskip]{92.0} & MiP & \cellbg[medgreen]{131} & \cellbg[medgreen]{53.8} & \cellbg[medgreen]{338} & \cellbg[deepgreen]{38.5} \\
& & & & & &  Well-defined & 148 & 0.3 & \cellbg[deepgreen]{305} & 11.5 \\
\midrule
\multirow{2}{*}[-0.5\baselineskip]{\textbf{Phi-3-medium-128k}} & \multirow{2}{*}[-0.5\baselineskip]{MiP} & \multirow{2}{*}[-0.5\baselineskip]{1465} & \multirow{2}{*}[-0.5\baselineskip]{48.0} & \multirow{2}{*}[-0.5\baselineskip]{125} & \multirow{2}{*}[-0.5\baselineskip]{\cellbg[lightgreen]{98.7}} & MiP & 210 & \cellbg[lightgreen]{47.8} & \cellbg[medgreen]{427} & 23.1 \\
& & & & & &  Well-defined & 216 & 1.0 & 1549 & 3.8 \\
\arrayrulecolor{black}\midrule
\multicolumn{11}{c}{\textbf{Reasoning Models}} \\ 
\midrule
\multirow{2}{*}[-0.5\baselineskip]{\textbf{GPT-o1}} & \multirow{2}{*}[-0.5\baselineskip]{MiP} & \multirow{2}{*}[-0.5\baselineskip]{1123} & \multirow{2}{*}[-0.5\baselineskip]{\cellbg[medgreen]{78.0}} & \multirow{2}{*}[-0.5\baselineskip]{581} & \multirow{2}{*}[-0.5\baselineskip]{\cellbg[medgreen]{99.0}} & MiP & 838 & 55.7 & 4189 & \cellbg[medgreen]{30.8} \\
& & & & & &  Well-defined & 348 & 0.3 & 2502 & 0.0 \\
\arrayrulecolor{gray}
\midrule
\multirow{2}{*}[-0.5\baselineskip]{\textbf{GPT-o1mini}} & \multirow{2}{*}[-0.5\baselineskip]{MiP} & \multirow{2}{*}[-0.5\baselineskip]{958} & \multirow{2}{*}[-0.5\baselineskip]{66.0} & \multirow{2}{*}[-0.5\baselineskip]{639} & \multirow{2}{*}[-0.5\baselineskip]{96.7} & MiP & 762 & 40.0 & 2193 & \cellbg[lightgreen]{25.0} \\
& & & & & &  Well-defined & 449 & 1.2 & 1913 & 0.0 \\
\midrule
\multirow{2}{*}[-0.5\baselineskip]{\textbf{GPT-o3mini}} & \multirow{2}{*}[-0.5\baselineskip]{MiP} & \multirow{2}{*}[-0.5\baselineskip]{1025} & \multirow{2}{*}[-0.5\baselineskip]{\cellbg[lightgreen]{76.0}} & \multirow{2}{*}[-0.5\baselineskip]{{1299}} & \multirow{2}{*}[-0.5\baselineskip]{93.0} & MiP & 1516 & {23.7} & 3772 & 11.5 \\
& & & & & &  Well-defined & 384 & 1.4 & 1553 & 0.0 \\
\midrule
\multirow{2}{*}[-0.5\baselineskip]{\textbf{DS Distill Qwen2.5-32B}} & \multirow{2}{*}[-0.5\baselineskip]{MiP} & \multirow{2}{*}[-0.5\baselineskip]{\cellbg[deepred]{12911}} & \multirow{2}{*}[-0.5\baselineskip]{42.0} & \multirow{2}{*}[-0.5\baselineskip]{921} & \multirow{2}{*}[-0.5\baselineskip]{{88.3}} & MiP & {2302} & 24.6 & \cellbg[medred]{9876} & \cellbg[lightred]{5.8} \\
& & & & & &  Well-defined & 519 & 0.2 & 3246 & 0.0 \\
\midrule
\multirow{2}{*}[-0.5\baselineskip]{\textbf{DeepSeek R1}} & \multirow{2}{*}[-0.5\baselineskip]{MiP} & \multirow{2}{*}[-0.5\baselineskip]{{4757}} & \multirow{2}{*}[-0.5\baselineskip]{\cellbg[medred]{6.0}} & \multirow{2}{*}[-0.5\baselineskip]{\cellbg[lightred]{1996}} & \multirow{2}{*}[-0.5\baselineskip]{\cellbg[lightred]{84.3}} & MiP & \cellbg[lightred]{3246} & \cellbg[lightred]{16.5} & {7268} & \cellbg[medred]{3.8} \\
& & & & & &  Well-defined & 1226 & 0.2 & 3200 & 1.9 \\
\midrule
\multirow{2}{*}[-0.5\baselineskip]{\textbf{S1.1-32B}} & \multirow{2}{*}[-0.5\baselineskip]{MiP} & \multirow{2}{*}[-0.5\baselineskip]{\cellbg[lightred]{5284}} & \multirow{2}{*}[-0.5\baselineskip]{\cellbg[lightred]{18.0}} & \multirow{2}{*}[-0.5\baselineskip]{\cellbg[medred]{3358}} & \multirow{2}{*}[-0.5\baselineskip]{\cellbg[medred]{57.0}} & MiP & \cellbg[deepred]{5024} & \cellbg[medred]{16.0} & \cellbg[lightred]{9322} & {15.4} \\
& & & & & &  Well-defined & 1896 & 0.2 & 5037 & 0.0 \\
\midrule
\multirow{2}{*}[-0.5\baselineskip]{\textbf{QwQ-32B}} & \multirow{2}{*}[-0.5\baselineskip]{MiP} & \multirow{2}{*}[-0.5\baselineskip]{\cellbg[medred]{7937}} & \multirow{2}{*}[-0.5\baselineskip]{\cellbg[deepred]{0.0}} & \multirow{2}{*}[-0.5\baselineskip]{\cellbg[deepred]{3487}} & \multirow{2}{*}[-0.5\baselineskip]{\cellbg[deepred]{56.3}} & MiP & \cellbg[medred]{4780} & \cellbg[deepred]{10.1} & \cellbg[deepred]{10242} & \cellbg[deepred]{1.9} \\
& & & & & &  Well-defined & 1896 & 0.2 & 5037 & 0.0 \\

\arrayrulecolor{black}\bottomrule
\end{tabular}
}
\caption{Comparing response length and abstain rate across different MiP datasets. Shorter lengths and higher abstain rates are preferred. For each column, the top-3 preferred values are colored in green, otherwise red. \textbf{MiP-Overthinking, reflected by longer response with low abstain rate, is commonly observed on most existing reasoning models across all datasets, indicating a critical drawback of existing reasoning models. }}
\label{tab:main_tab}
\end{table*}

Moreover, Table \ref{tab:main_tab} further presents the comparisons on length and abstain rate on other MiP datasets we curated. 
The preferred results are colored green (shorter responses and higher abstain rate for MiP questions), and the worse results are colored red, from which we can easily discover that reasoning models are prone to generate long responses while having low abstain rates across all datasets, indicating the consistent MiP Overthinking issue of existing reasoning models. 
In addition, by comparing the behaviors of models on different datasets, we can observe that for the relatively harder dataset (MiP-MATH), all models generate relatively longer responses and obtain lower abstain rates, indicating that harder MiP questions require reasoning capabilities.

\subsection{Thinking Patterns through Tokens}
\label{sec:token_cnt}

\begin{table*}[t]
\centering
\renewcommand\arraystretch{1.0}
\resizebox{\textwidth}{!}{%
\begin{tabular}{l|c|cc|cc|cc|cc|cc|cc}
\toprule
\multirow{2}{*}{\textbf{Models}} & \multirow{2}{*}{\textbf{Type}} & \multicolumn{2}{c|}{\textbf{Alternatively}} & \multicolumn{2}{c|}{\textbf{Wait}} & \multicolumn{2}{c|}{\textbf{Check}} & \multicolumn{2}{c|}{\textbf{But}} & \multicolumn{2}{c|}{\textbf{Hypothesis}} & \multicolumn{2}{c}{\textbf{Step}} \\
& & Cnt.  & $\Delta$ & Cnt.  & $\Delta$ & Cnt.  & $\Delta$ & Cnt.  & $\Delta$ & Cnt.  & $\Delta$ & Cnt.  & $\Delta$ \\
\midrule
\multicolumn{14}{c}{\textbf{Non-Reasoning Models}} \\ \midrule
\multirow{2}{*}[-0.5\baselineskip]{\textbf{Qwen2.5-32B}} & MiP & 0.0 & \multirow{2}{*}{0.0} & 0.0 & \multirow{2}{*}{0.0} & 0.0 & \multirow{2}{*}{0.0} & 0.3 & \multirow{2}{*}{0.2} & 0.0 & \multirow{2}{*}{0.0} & 4.3 & \multirow{2}{*}{\cellbg[medgreen]{-1.3}} \\
&  Well-defined & 0.0 &  & 0.0 &  & 0.0 &  & 0.1 &  & 0.0 &  & 5.6 &  \\
\arrayrulecolor{gray}\midrule
\multirow{2}{*}[-0.5\baselineskip]{\textbf{GPT-4o}} & MiP & 0.0 & \multirow{2}{*}{0.0} & 0.0 & \multirow{2}{*}{0.0} & 0.0 & \multirow{2}{*}{0.0} & 0.3 & \multirow{2}{*}{0.2} & 0.0 & \multirow{2}{*}{0.0} & 4.7 & \multirow{2}{*}{\cellbg[medgreen]{-1.5}} \\
&  Well-defined & 0.0 &  & 0.0 &  & 0.0 &  & 0.1 &  & 0.0 &  & 6.2 &  \\
\midrule
\multirow{2}{*}[-0.5\baselineskip]{\textbf{Gemini 1.5}} & MiP & 0.0 & \multirow{2}{*}{0.0} & 0.0 & \multirow{2}{*}{0.0} & 0.0 & \multirow{2}{*}{0.0} & 0.1 & \multirow{2}{*}{0.1} & 0.0 & \multirow{2}{*}{0.0} & 1.6 & \multirow{2}{*}{\cellbg[medgreen]{-2.2}} \\
&  Well-defined & 0.0 &  & 0.0 &  & 0.0 &  & 0.0 &  & 0.0 &  & 3.8 &  \\
\midrule
\multirow{2}{*}[-0.5\baselineskip]{\textbf{Gemma-2-27B}} & MiP & 0.0 & \multirow{2}{*}{0.0} & 0.0 & \multirow{2}{*}{0.0} & 0.0 & \multirow{2}{*}{0.0} & 0.1 & \multirow{2}{*}{0.1} & 0.0 & \multirow{2}{*}{0.0} & 5.2 & \multirow{2}{*}{\cellbg[medgreen]{-0.5}} \\
&  Well-defined & 0.0 &  & 0.0 &  & 0.0 &  & 0.0 &  & 0.0 &  & 5.7 &  \\
\midrule
\multicolumn{14}{c}{\textbf{Reasoning Models}} \\ \midrule
\multirow{2}{*}[-0.5\baselineskip]{\textbf{DS-Distill Qwen}} & MiP & 11.5 & \multirow{2}{*}{11.4} & 19.7 & \multirow{2}{*}{19.3} & 1.0 & \multirow{2}{*}{0.8} & 40.1 & \multirow{2}{*}{39.3} & 38.4 & \multirow{2}{*}{38.0} & 54.9 & \multirow{2}{*}{\cellbg[medred]{42.2}} \\
&  Well-defined & 0.1 &  & 0.4 &  & 0.2 &  & 0.8 &  & 0.4 &  & 12.7 &  \\
\midrule
\multirow{2}{*}[-0.5\baselineskip]{\textbf{DeepSeek R1}} & MiP & 16.9 & \multirow{2}{*}{15.2} & 14.4 & \multirow{2}{*}{10.9} & 3.8 & \multirow{2}{*}{1.3} & 49.4 & \multirow{2}{*}{42.1} & 44.7 & \multirow{2}{*}{40.4} & 54.2 & \multirow{2}{*}{\cellbg[medred]{33.0}} \\
&  Well-defined & 1.7 &  & 3.5 &  & 2.5 &  & 7.3 &  & 4.3 &  & 21.2 &  \\
\midrule
\multirow{2}{*}[-0.5\baselineskip]{\textbf{S1.1}} & MiP & 42.0 & \multirow{2}{*}{38.0} & 21.9 & \multirow{2}{*}{15.9} & 5.5 & \multirow{2}{*}{2.5} & 87.2 & \multirow{2}{*}{74.1} & 84.8 & \multirow{2}{*}{77.0} & 79.9 & \multirow{2}{*}{\cellbg[medred]{50.9}} \\
&  Well-defined & 4.0 &  & 6.0 &  & 3.0 &  & 13.1 &  & 7.8 &  & 29.0 &  \\
\midrule
\multirow{2}{*}[-0.5\baselineskip]{\textbf{QwQ}} & MiP & 47.0 & \multirow{2}{*}{40.3} & 19.4 & \multirow{2}{*}{13.0} & 5.0 & \multirow{2}{*}{1.6} & 66.1 & \multirow{2}{*}{54.2} & 94.1 & \multirow{2}{*}{81.7} & 97.9 & \multirow{2}{*}{\cellbg[medred]{58.7}} \\
&  Well-defined & 6.7 &  & 6.4 &  & 3.4 &  & 11.9 &  & 12.4 &  & 39.2 &  \\
\arrayrulecolor{black}\bottomrule
\end{tabular}
}
\caption{Comparisons of reasoning-related token counts on MiP-GSM8K dataset. 
\textit{Hypothesis} category includes several key words, including \textit{perhaps}, \textit{maybe}, and \textit{might}.
\textit{Step} represents the step counts, spited by \textit{\textbackslash n\textbackslash n}, where negative values are colored in green and positive in red. 
$\Delta$ denotes the difference between MiP and well-defined questions.
\textbf{When facing MiP questions, reasoning models encounter explosive growths on reasoning-related tokens and steps, indicating a severe abuse of thinking patterns, while non-reasoning models use fewer steps for MiP questions than well-defined ones.}  
}
\label{tab:wordcount}
\end{table*}


To gain deeper insight into the MiP-Overthinking issue, we compare the reasoning-related token distribution on the MiP-GSM8K dataset. As shown in Table \ref{tab:wordcount}, we break down the average usages of several token patterns related to the thinking process, as well as the number of steps for each model to solve the given questions. 
Specifically, values of \textit{alternatively}, \textit{wait}, \textit{check}, and \textit{but} can be directly counted from the model responses, including the thinking paths of reasoning models. 
\textit{Hypothesis} category includes several key words, including \textit{perhaps}, \textit{maybe}, and \textit{might}.
\textit{Step} represents the step counts, spited by \textit{\textbackslash n\textbackslash n}. 

Reasoning models exhibit much higher occurrence of tokens such as \textit{alternatively}, \textit{wait}, and \textit{check}, compared with non-reasoning models, whose frequencies remain close to zero, indicating their advanced thinking capabilities. 
However, when moving from well-defined to MiP questions, reasoning models encounter explosive growths on reasoning-related tokens, indicating a large redundancy in thinking patterns. 
Moreover, when comparing the changes of steps, reasoning models exhibit a large increase in step count for MiP questions, while non-reasoning models typically show fewer steps, suggesting they quickly conclude the question is unanswerable. 
With this gap, together with the consistently better abstain rates of the non-reasoning models, we conclude that \textbf{the lengthy reasoning steps are mostly redundant and indicate self-doubt thinking patterns for reasoning models.}  


\clearpage
\subsection{Step-level Similarities}
\label{sec:sentence_sim}


\begin{wrapfigure}[22]{r}{0.5\textwidth}
    \vspace{-15pt} 
    \centering
    \includegraphics[width=0.5\textwidth]{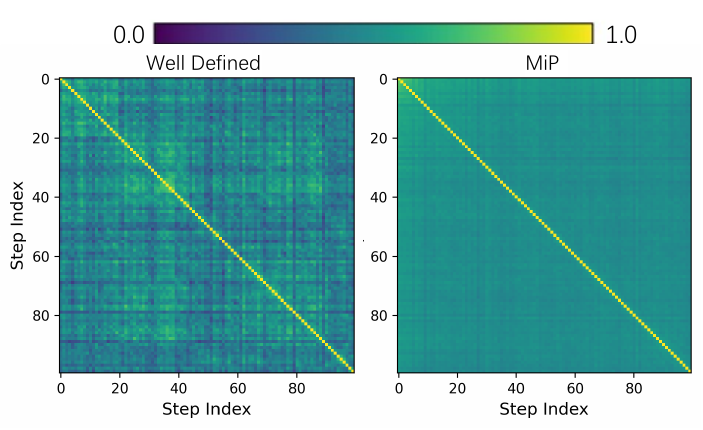}
    \vspace{-20pt} 
    \captionsetup{width=0.98\linewidth}
    \caption{The step-level similarity heatmaps for s1.1 responses towards well-defined (left) and MiP (right) questions in MiP-GSM8K dataset. To avoid differences in matrix size, we only consider responses with more than 50 steps and visualize the average simialrity matrix across first 50 steps. \textbf{The heatmap for MiP questions has a higher averaged similarity and lower standard variance, also shown in the heatmap, which indicates the considerable redundancy in its content when responding to MiP questions. }} 
    \label{fig:similarity}
\end{wrapfigure}

To further assess how redundant the generated content becomes under MiP conditions, we examine the step-level similarity within the model’s responses on our MiP-GSM8K dataset. Specifically, we divide each response into discrete steps, split by \textit{\textbackslash n\textbackslash n}, and compute pairwise cosine similarity scores with embeddings generated by ``all-MiniLM-L6-v2'' \citep{reimers-gurevych-2019-sentence}. The visualization is shown in Figure \ref{fig:similarity}, where each value in the heatmap metrix represents the averaged cosine similarities between the corresponding step index. The average similarity score for well-defined question is \textbf{0.45} and \textbf{0.50} for MiP response. The variance is \textbf{7.9e-3} and \textbf{8.2e-4} respectively.


As shown in the figure, responses to MiP questions have greater overall similarity across steps and lower standard variance, indicating the considerable redundancy in the content. 
This means, in many instances, \textbf{the model revisits similar partial reasoning or repeats previous sentences with only minor changes, showing a potential self-trapping issue.} 
Together, these patterns confirm that MiP questions induce a high degree of repetitive content in reasoning models. Rather than terminating early to conclude for insufficient premise, the models fill their reasoning paths with repetitive re-checks and reiterations, significantly inflating token usage without improving real abstain rates.

\subsection{Thinking Patterns through Example}

\begin{figure*}[tbh]
    \centering
    \includegraphics[width=0.88\linewidth]{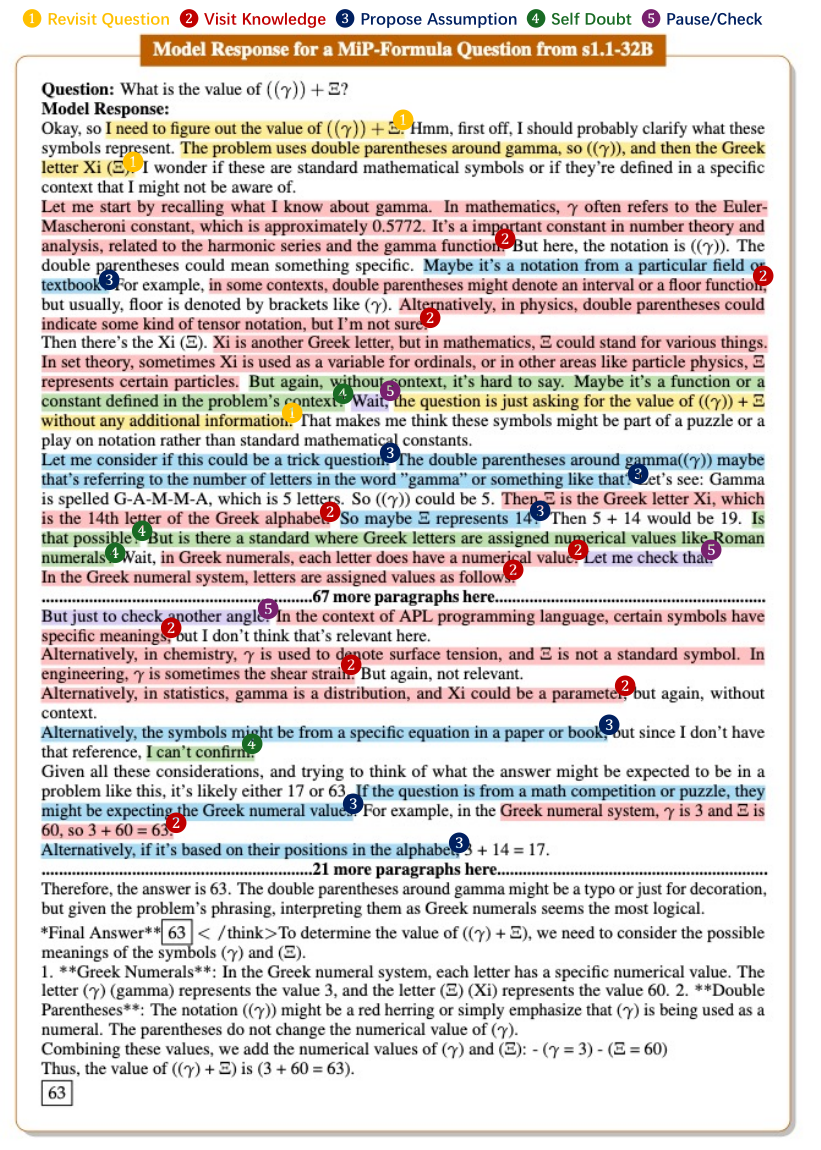}
    \caption{An example of reasoning model (s1.1-32B) response to a MiP question. The response exhibits five distinct thinking patterns, highlighted in different colors:  \textcolor{Dandelion}{\textbf{\textcircled{\raisebox{-0.9pt}{1}}Revisit Question}} (yellow), where the model reexamines the original query;  \textcolor{OrangeRed}{\textbf{\textcircled{\raisebox{-0.9pt}{2}}Visit Knowledge}} (red), where the model accesses domain-specific knowledge; \textcolor{RoyalBlue}{\textbf{\textcircled{\raisebox{-0.9pt}{3}}Propose Assumption}} (blue), where the model proposes and investigates various hypotheses;  \textcolor{OliveGreen}{\textbf{\textcircled{\raisebox{-0.9pt}{4}}Self Doubt}} (green), where the model questions its own reasoning and expresses uncertainty; and \textcolor{Fuchsia}{\textbf{\textcircled{\raisebox{-0.9pt}{5}}Pause/Check}} (purple), where the model pauses to review previous steps. These patterns demonstrate the model's complex but potentially inefficient reasoning process when confronted with missing premises.}
    \label{fig:example}
    \vspace{-5mm}
\end{figure*}

To further understand what happens in the reasoning chain of reasoning models when faced an ill-post input, we present an example of reasoning model's response to a MiP question in Figure~\ref{fig:example}. We summarize five major thinking patterns we found in the example and highlight them with different colors. We can observe from the example that the model abuses these patterns to generate long responses, while the responses are not only redundant but also not helpful for the model to abstain from the given MiP question. More examples can be found in the appendix~\ref{sec:example}.

\vspace{-10pt}
\section{Further Discussion}

\subsection{Do Models know premises are missing?}
To investigate whether reasoning models recognize the potential unsolvability of questions during their reasoning process, we conducted a detailed analysis of their reasoning chains. We segmented each reasoning chain into discrete steps using \textit{\textbackslash n\textbackslash n} as delimiters and performed step-wise verification to detect whether models express doubt on the question solvability. We introduce two key metrics for this analysis: \textbf{In-Process Suspicion Rate}, which measures the percentage of responses where the model expresses doubt about solvability during reasoning, and \textbf{First Suspicion Index}, which captures the average step number at which the model first suspects the missing premise.
To ensure robust evaluation, we employed GPT-4o to assess each step three times, using majority voting for our final step-level result. The quantitative results of this analysis are presented in Table \ref{tab:occur}.

\begin{table}[t]
    \centering
    \renewcommand\arraystretch{0.7}
    \resizebox{\textwidth}{!}{%
    \begin{tabular}{l|cccc|cccc}
        \toprule
        & \multicolumn{4}{c|}{MiP-Formula} & \multicolumn{4}{c}{MiP-GSMR} \\
        \cline{2-5} \cline{6-9}
        Model & DeepSeek-R1 & DS-Qwen & QwQ & S1.1 & DeepSeek-R1 & DS-Qwen & QwQ & S1.1 \\ \midrule
        In-Process Suspicion Rate
 & 100\% & 100\% & 100\% & 100\% & 95.5\% & 83.3\% & 99.6\% & 100\% \\ \arrayrulecolor{gray}
       In-Process First Suspicion Index
 & 1.32 & 1.36 & 1.42 & 1.16 & 2.01 & 3.90  & 1.77 & 1.61 \\ 
        \arrayrulecolor{black}\bottomrule
    \end{tabular}}
    \caption{The in-process insufficiency suspicion information across different reasoning models on MiP-Formula and MiP-GSMR datasets. The in-process insufficiency suspicion is defined as when the reasoning model suspects the given question is unsolvable during its thinking process. \textit{In-Process Suspicion Rate} represents how many percent of the samples trigger the in-process suspicion. \textit{First Suspicion Index} is the averaged step index where the model first suspects the question's validity. \textbf{Most reasoning models do notice the existence of MiP at the very early steps, but they still suffer from low abstain rate and cannot confidently stop the thinking. }
    \vspace{-15pt}
    }
    \label{tab:occur}
\end{table}

\begin{figure*}[h]
    \centering
    \includegraphics[width=0.9\linewidth]{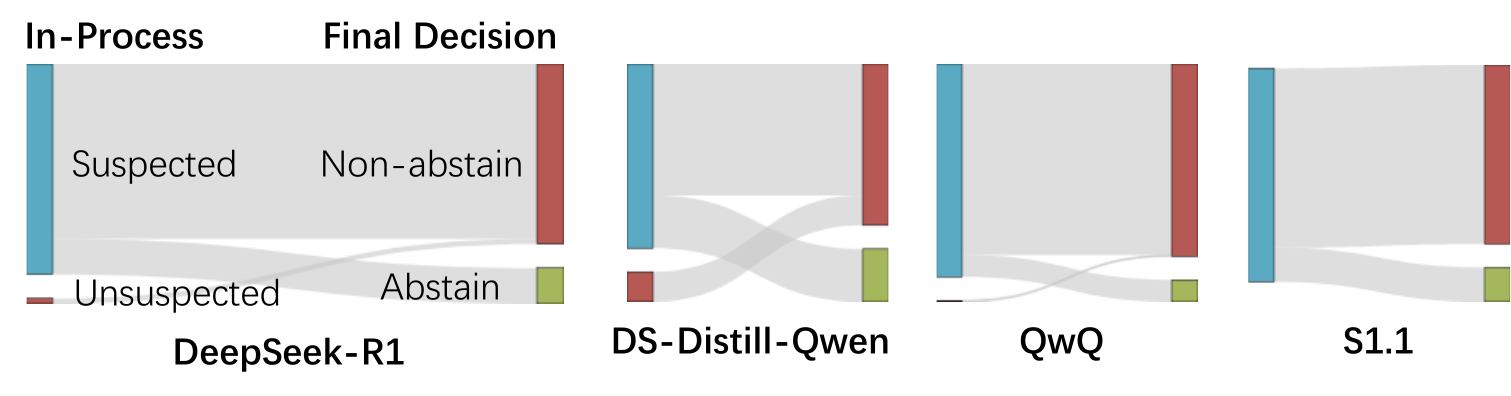}
    \vspace{-4mm}
    \caption{The transition flow between in-process suspicion of MiP and the final successful abstention on different reasoning models. For each Sankey diagram, the left bars represent whether the model suspects the given question is unsolvable during its thinking process, i.e., \textit{Suspected} or \textit{Unsuspected}; the right bars represent the final abstention, categorized into \textit{Abstain} (preferred) or \textit{Non-abstain}. \textbf{Most existing reasoning models have suspected that the given question might be unsolvable, but only for a very small portion, the models insist on their suspicion. }
    }
    \label{fig:occur}
    \vspace{-4mm}
\end{figure*}

As we can see from the table, most of the existing reasoning models have suspected that the given question might be unsolvable at the very early stage of their reasoning process, demonstrating the ability of reasoning models to recognize the potential MiP. 
However, these reasoning models lack critical thinking capabilities: they are prone to keep digging the given unsolvable question by re-visiting the question and related definitions again and again and again, rather than question the solvability of the given question. 
Thus, as visualized in Figure \ref{fig:occur}, despite existing reasoning models suspecting the solvability of most of the given MiP questions, they only abstain a very small proportion of them.  

Based on the above observations, we conclude that reasoning models actually have the capabilities to find out that the given MiP question is not solvable, but they do not ``dare'' to abstain it. Thus, our MiP-Overthinking issue indicates the lack of critical thinking abilities of reasoning models.  

\vspace{-4mm}


\vspace{-25pt}
\subsection{What Caused MiP-Overthinking?}

Figure~\ref{fig:mian_fig} demonstrates that MiP-Overthinking manifests across both RL-based and SFT-based reasoning models. We hypothesize this phenomenon primarily originates from inadequate length constraints during the rule-based reinforcement learning phase of RL-based models, subsequently propagating to SFT-based models through distillation.

\begin{wrapfigure}[20]{r}{0.45\textwidth}
    \vspace{-10pt} 
    \centering
    \includegraphics[width=0.5\textwidth]{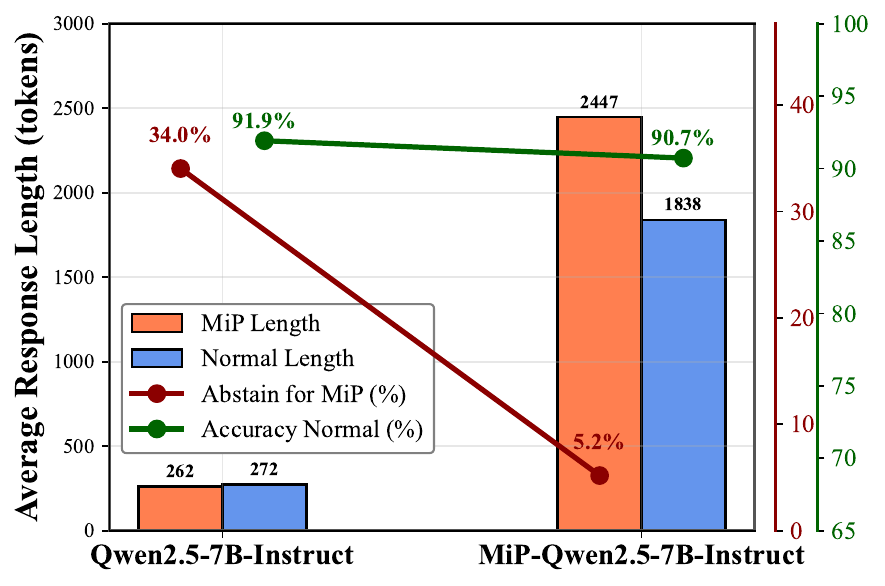}
    \vspace{-10pt} 
    \captionsetup{width=0.98\linewidth}
    \caption{Comparison of response length, abstain rate of MiP, and accuracy of well-defined questions before and after tuning on 50 responses from DeepSeek-R1 on the MiP-Formula dataset. The results demonstrate rapid onset of MiP-Overthinking behavior after exposure to a small number of MiP examples during fine-tuning.} 
    \label{fig:sft_result}
\end{wrapfigure}
\vspace{-5pt}

Current RL-based reasoning models predominantly employ rule-based training focused on format and accuracy rewards~\citep{shao2024deepseekmathpushinglimitsmathematical, sui2025stopoverthinkingsurveyefficient}, with some incorporating step or length rewards to promote thorough reasoning~\citep{openr1}. This approach can lead to reward hacking, where models explore excessive reasoning patterns to achieve correct answers~\citep{aggarwal2025l1controllinglongreasoning, shen2025dastdifficultyadaptiveslowthinkinglarge, luo2025o1prunerlengthharmonizingfinetuningo1like}. 

To demonstrate the transmissibility of this behavior through distillation \citep{xu2024surveyknowledgedistillationlarge}, we finetune Qwen-2.5-7B-Instruct using small-scale $50$ MiP responses generated by DeepSeek-R1 on the MiP-Formula dataset. As shown in Figure~\ref{fig:sft_result}, the fine-tuned model exhibits clear MiP-Overthinking characteristics when evaluated on GSM8K: significantly increased response lengths for both MiP and well-defined questions, emergence of a length disparity between MiP and well-defined responses previously absent in the original model, and decreased abstain rates.

\section{Related Work}
\label{sec:related}
\subsection{Reasoning Large Language Model}
Recent advances in Large Language Models (LLMs) have sparked significant research interest in enhancing their reasoning capabilities~\citep{ahn-etal-2024-large, besta2025reasoninglanguagemodelsblueprint, chen2025reasoningerasurveylong}. 
Research has focused on improving these capabilities through various post-training approaches. Several studies have employed reinforcement learning techniques to guide models toward more effective reasoning strategies~\citep{shao2024deepseekmathpushinglimitsmathematical, xiong2025selfrewardingcorrectionmathematicalreasoning, cui2025processreinforcementimplicitrewards}. Additionally, researchers have demonstrated that instruction tuning on carefully curated, high-quality datasets can significantly enhance reasoning performance~\citep{ye2025limoreasoning, muennighoff2025s1simpletesttimescaling}.

While Reasoning Models have demonstrated impressive performance on various benchmarks, recent studies have begun to critically examine the quality and efficiency of their reasoning processes. \citet{xia2025evaluatingmathematicalreasoningaccuracy} conducted a comprehensive analysis of RLMs' reasoning quality, revealing significant redundancy in their solution approaches. Further investigations~\citep{chen2025think23overthinkingo1like, cuadron2025dangeroverthinkingexaminingreasoningaction, qu2025surveyefficientreasoninglarge, liu2025efficientinferencelargereasoning} identified a concerning "overthinking" phenomenon, where reasoning model generate unnecessarily verbose solutions even for simple problems. Building on these observations, \citet{kumar2025overthinkslowdownattacksreasoning} demonstrated the potential security implications of this behavior by developing a slowdown attack that exploits overthinking through input perturbation.

\subsection{Test-time Scaling}
In contrast to earlier research on training-time scaling laws~\citep{kaplan2020scalinglawsneurallanguage}, recent literature has increasingly focused on test-time performance scaling strategies, which aim to enhance model performance by optimizing inference-time token generation~\citep{snell2024scalingllmtesttimecompute, learningtoreason}. These approaches can be categorized into several primary methodologies: parallel sampling techniques~\citep{brown2024largelanguagemonkeysscaling, levi2024simplemodelinferencescaling}, which generate multiple candidate responses and select the optimal output; sequential refinement approaches~\citep{snell2024scalingllmtesttimecompute, lee2025evolvingdeeperllmthinking}, which enable iterative improvement of previous outputs; and tree-based methods~\citep{gandhi2024streamsearchsoslearning, hou2025advancinglanguagemodelreasoning}, which combine elements of both parallel and sequential approaches. While the prevailing consensus suggests that increased token generation during inference enhances reasoning capabilities, our investigation reveals a concerning counterpoint: under certain conditions, extended responses can lead to computational inefficiency and, paradoxically, degraded performance outcomes.

\subsection{Models' Behavior Study in Ambiguous Condition} 

LLMs are prone to hallucination~\citep{Huang_2025,xu2025hallucinationinevitableinnatelimitation}, generating non-existent conditions that compromise trustworthiness. An essential aspect of reliability is the ability to abstain under uncertainty. Prior work~\citep{cole2023selectivelyansweringambiguousquestions, amayuelas2024knowledgeknowledgeexploringknownunknowns, zhou2023navigatinggreyareaexpressions} has proposed benchmarks assessing LLMs' recognition of knowledge limits when facing ambiguous or challenging queries. Different from theirs, our study explores reasoning models under MiP condition. Surprisingly, we find these specialized models exhibit prolonged reasoning and inferior performance.


\vspace{-5pt}
\section{Conclusion}
\vspace{-5pt}
We introduce the Overthinking under Missing Premise (MiP-Overthinking) issue, which is a widespread but still under-explored phenomenon for current reasoning models. 
In this phenomenon, when faced with ill-defined unsolvable questions with missing premises, existing models generate dramatically long responses while having very low abstain rates. 
With systematic investigation of this phenomenon, our findings show that while these models sometimes suspect the given MiP question is not solvable in the early state of the thinking process, they typically fail to act on those suspicions and instead generating repetitive and redundant thinking traces with the final answer that does not address the missing premises, indicating a lack of critical thinking capability.  
This behavior highlights a pressing gap: current training recipes for reasoning models, which emphasize thorough chains of thought, do not sufficiently reward critical thinking or early exit from unsolvable tasks.



\bibliography{colm2025_conference}
\bibliographystyle{colm2025_conference}

\clearpage
\appendix

\startcontents[appendix]
\printcontents[appendix]{ }{0}{\section*{Table of Contents for Appendix}}

\clearpage
\section{Detailed Experimental Setup}
\label{sec:experiment}
\subsection{Models}
We leverage a series of non-reasoning and reasoning model for our study, from both open-source and proprietary source with different training recipes. The non-reasoning models we use include Qwen2.5-32B-Instruct~\cite{qwen2.5}, Gemma-2-27B-it~\cite{gemmateam2024gemma2improvingopen}, Phi-3-medium-128k~\cite{abdin2024phi3technicalreporthighly} ,GPT-4o~\cite{openai2024gpt4technicalreport} and Gemini1.5~\cite{geminiteam2024gemini15unlockingmultimodal}. The reasoning models we use are QwQ-32B~\cite{qwq32b}, DeepSeek-R1-Distill-Qwen-32B~\cite{deepseekai2025deepseekr1incentivizingreasoningcapability}, S1.1~\cite{muennighoff2025s1simpletesttimescaling}, DeepSeek-R1~\cite{deepseekai2025deepseekr1incentivizingreasoningcapability}, GPT-o1~\cite{openai2024o1}, GPT-o1mini~\cite{openai2024o1mini} and GPT-o3mini~\cite{openai2025o3mini}.

\subsection{Evaluation Metrics}
In Section~\ref{sec:main}, we measure response length by considering both reasoning and answer components. For open-source models, we employ model-specific tokenizers to calculate token counts, while for proprietary models, we obtain generation lengths via their APIs. To determine abstain rates, we parse responses by paragraphs (delimited by `\textbackslash n\textbackslash n`) and analyze the final two paragraphs as the model's conclusion. These conclusions, along with reference answers when available, are evaluated by GPT-4o to assess whether the model provides a definitive answer or abstains. For data sets with reference answers (GSM8K and MATH), GPT-4o also evaluates the correctness of the response. The prompt we use for evaluation can be found in Appendix~\ref{sec:prompt}.

\subsection{Generation Setting}
For all open-source models, we employ greedy decoding and utilize the default chat template specific to each model. We deliberately omit system prompts prior to posing questions to maintain consistency across evaluations. For proprietary models, we adhere to their default parameter configurations as provided by their respective APIs. In the case of GPT-o1mini and GPT-o3mini, we configure the `reasoning\_effort' parameter to the medium setting by default.

\section{Data Construction Details}
\label{sec:data_detail}

To systematically investigate this MiP-Overthinking issue, we construct a suite of MiP questions in a controllable manner. Our MiP questions are sourced from $3$ math datasets across different qualities, including SVAMP, GSM8K, and MATH 500. In addition, we also construct a synthetic dataset, rule-based Formula, for evaluation. 

\header{MiP-Formula}
We construct a dataset of $50$ synthetic unsolvable formulas in a rule-based manner. The formulas are generated recursively through a combination of variables and operators, with a maximum recursion depth of three. The variable set comprises numerical values, Latin letters, and Greek symbols. The operator set includes arithmetic operators ('$+$', '$-$'), set operators ('$\cup$', '$\supset$'), mathematical functions ('sin', 'sqrt'), and construct operators ('$\sum$', '$\nabla$'). To ensure the formulas are fundamentally unsolvable, we enforce the inclusion of at least one unassigned variable in each formula, excluding commonly recognized mathematical or physical constants such as '$e$', '$\pi$', and '$g$'. While these formulas may appear complex at a glance, their unsolvability should be immediately apparent due to the presence of undefined variables.

\header{MiP-SVAMP}
We utilize SVAMP~\citep{patel-etal-2021-nlp}, a benchmark dataset comprising $1,000$ elementary-school-level mathematical word problems, where each instance consists of a problem body and an associated question. 
The MiP questions can be generated by randomly permuting the problem bodies and associated questions. 
To maintain dataset integrity, we manually select $300$ permuted questions after a thorough human evaluation to eliminate any inadvertently solvable questions that may exist. 
The resulting problems contain clear logical inconsistencies between their body and question components, making their unsolvability readily apparent without additional context.

\header{MiP-GSM8K}
We further utilize GSM8K~\citep{cobbe2021gsm8k}, a grade school mathematics dataset that presents more complex challenges compared to SVAMP. The questions in GSM8K typically contain multiple numerical conditions and require certain reasoning capabilities to arrive at solutions. 
The MiP question can be constructed by randomly removing a necessary premise from the original solvable question. 
We first identify the questions containing two or three numerical conditions and then randomly eliminate one numerical condition per question.
Subsequently, a thorough human verification is conducted to filter out those questions that are still solvable in some way and finally obtain $582$ MiP questions. 
Compared with previous MiP questions, questions from this source require the basic logical analysis of models to identify that the question is unsolvable.  

\header{MiP-MATH}
For the MATH dataset~\citep{hendrycksmath2021}, which comprises challenging competition-level mathematical questions, it is hard to build a rule-based filtering mechanism before human evaluation. Thus, we directly read through all the questions in MATH500 and manually select $58$ questions that are feasible for constructing the MiP questions and remove one necessary premise from the question. 
Due to the sophisticated nature of this data source, identifying the insufficiency of these instances requires substantial mathematical reasoning capabilities, testing models' ability to recognize unsolvability in complex mathematical contexts.

\section{Prompt Template for Evaluation}
\label{sec:prompt}

As we need LLM-as-a-judge to evaluate the open-end generations of the models in various experiment in this study, in this section we showcase the prompt template we use for each kind of evaluation.

For the evaluation of the models' answer accuracy and abstain rate, we adopt the following prompt templates designed for 'paired' and 'non-paired' data, respectively. As we observe that some models, for example Gemma-2-27B-IT, often output an additional \textit{\textbackslash n\textbackslash n} at the end of response, we take the last two paragraph segmented by \textit{\textbackslash n\textbackslash n} to avoid pasing in an empty string.

\begin{figure*}[!h]
\begin{tcolorbox}[
  enhanced, 
  colframe=teal!75!black, 
  colback=white, 
  coltitle=white, 
  colbacktitle=teal!75!black, 
  width=\linewidth, 
  arc=2mm, 
  auto outer arc, 
  boxrule=0.5pt, 
  left=10pt, 
  right=10pt, 
  drop shadow={black!50!white},
  top=10pt, 
  bottom=10pt, 
  title=\textbf{Prompt Template for Response Evaluation of Formula and SVAMP}, 
  fonttitle=\bfseries, 
  title code={\node[rounded corners, fill=blue!75!black, draw=none, text=white] at (frame.title) {\textbf{xxx}};}, 
  attach boxed title to top center={yshift=-2mm}, 
  boxed title style={sharp corners, size=small}, 
]
\small
You are a helpful assistant that evaluates the quality of a model's answer. You will be given a question and a model's answer. You need to evaluate the correctness of the model's answer. If the answer explicitly says that the condition of the question is insufficient, you should return 0. If the model provides an answer that is a number or formula with variables, you should return 1. Please only return the number, no other text.

Model answer: [model\_answer\_short]

\end{tcolorbox}
\caption{The prompt we use to evaluate the accuracy and abstain rate of the model on Formula and SVAMP. [model\_answer\_short] is the last two paragraphs of the model answer and [reference\_answer] is the answer for the orginal dataset.}
\label{fig:promt_eval_1}
\end{figure*}

\vspace{-15pt}
\begin{figure*}[!h]
\begin{tcolorbox}[
  enhanced, 
  colframe=teal!75!black, 
  colback=white, 
  coltitle=white, 
  colbacktitle=teal!75!black, 
  width=\linewidth, 
  arc=2mm, 
  auto outer arc, 
  boxrule=0.5pt, 
  left=10pt, 
  right=10pt, 
  drop shadow={black!50!white},
  top=10pt, 
  bottom=10pt, 
  title=\textbf{Prompt Template for Response Evaluation of GSM8K and MATH}, 
  fonttitle=\bfseries, 
  title code={\node[rounded corners, fill=blue!75!black, draw=none, text=white] at (frame.title) {\textbf{xxx}};}, 
  attach boxed title to top center={yshift=-2mm}, 
  boxed title style={sharp corners, size=small}, 
]
\small
You are a helpful assistant that evaluates the quality of a model's answer. You will be given a question and a model's answer. You need to evaluate the correctness of the model's answer. If the model output says that the condition of the question is insufficient, you should return 0. Otherwise, if the model give a clear answer and matches the reference answer, you should return 1. If the model's answer does not match the reference answer, you should return 2. Please only return the number, no other text.

Model answer: [model\_answer\_short]

Reference answer: [reference\_answer]
\end{tcolorbox}
\caption{The prompt we use to evaluate the accuracy and abstain rate of the model on GSM8K and MATH. [model\_answer\_short] is the last two paragraphs of the model answer and [reference\_answer] is the answer for the orginal dataset.}
\label{fig:promt_eval_2}
\end{figure*}

We use the prompt template in Figure~\ref{fig:promt_suspect} to find the first paragraph that the model suspected a missing premise. We pass in the response sequentially by paragraph until the GPT-4o give a positive response. In practice we find it is not very stable, so we repeat this process for 3 times and use the median value.
\begin{figure*}[!h]
\begin{tcolorbox}[
  enhanced, 
  colframe=teal!75!black, 
  colback=white, 
  coltitle=white, 
  colbacktitle=teal!75!black, 
  width=\linewidth, 
  arc=2mm, 
  auto outer arc, 
  boxrule=0.5pt, 
  left=10pt, 
  right=10pt, 
  drop shadow={black!50!white},
  top=10pt, 
  bottom=10pt, 
  title=\textbf{Prompt Template for Response Evaluation of GSM8K and MATH}, 
  fonttitle=\bfseries, 
  title code={\node[rounded corners, fill=blue!75!black, draw=none, text=white] at (frame.title) {\textbf{xxx}};}, 
  attach boxed title to top center={yshift=-2mm}, 
  boxed title style={sharp corners, size=small}, 
]
\small
You are a helpful assistant. You will be given a paragraph which is part of the answer to a question. You need to identify if the paragraph doubt that the answer depends on some other unspecified condition. 

Paragraph: [paragraph]

Answer in one word, yes or no.

\end{tcolorbox}
\caption{The prompt we use to judge if the model suspect there is a missing premise in the response paragraph. [paragraph] is the part of the model response spited by \textit{\textbackslash n\textbackslash n}}.
\label{fig:promt_suspect}
\end{figure*}

\section{Examples of Model Response}
\label{sec:example}

In this section, we present some examples of the model response of both non-reasoning and reasoning model on MiP data. As we can see from Figure~\ref{fig:example-1-1} and Figure~\ref{fig:example-2-1}, the non-reasoning models soon identify the missing premise issue of the question. They either abstain from answering the question, as in Figure~\ref{fig:example-1-1}, or friendly invite the user to provide more information. However, as we can see from Figure~\ref{fig:example-2-1} and Figure~\ref{fig:example-2-2}, reasoning models generate extremely verbose answers on these two apparently premise missing problems. What is worse, they fail to abstain to answer the question. The response in Figure~\ref{fig:example-2-1} arrives at an absurd answer, and the model in Figure~\ref{fig:example-2-2} generates a hallucinated answer based on its assumption rather than provided information.

\begin{figure*}[h]
\begin{tcolorbox}[
  enhanced, 
  colframe=cyan!75!black, 
  colback=white, 
  coltitle=white, 
  colbacktitle=cyan!75!black, 
  width=\linewidth, 
  arc=2mm, 
  auto outer arc, 
  boxrule=0.5pt, 
  left=10pt, 
  right=10pt, 
  drop shadow={black!50!white},
  top=10pt, 
  bottom=10pt, 
  title=\textbf{Model Response for a MiP-Formula Question from Gemini-1.5}, 
  fonttitle=\bfseries, 
  title code={ ode[rounded corners, fill=blue!75!black, draw=none, text=white] at (frame.title) {\textbf{xxx}};}, 
  attach boxed title to top center={yshift=-2mm}, 
  boxed title style={sharp corners, size=small}, 
]
\small
\textbf{Question:} What is the value of $((\gamma)) + \Xi$?

\textbf{Model Response:} 

The expression $((\gamma)) + \Xi$ involves two symbols that need to be defined to determine the value of the expression.\\$\gamma$ is the Euler-Mascheroni constant, approximately equal to 0.5772156649. The double parentheses notation $((\gamma))$ is not a standard mathematical notation. Without further definition of this notation, we cannot evaluate this term.\\$\Xi$ is the Riemann Xi function. It is related to the Riemann zeta function, but the notation $\Xi$ itself isn't a specific numerical value. The Riemann Xi function is defined as a function of a complex variable, not a constant.  So $\Xi$ by itself doesn't have a specific numerical value unless a specific value is substituted for its argument.\\Therefore, we cannot determine the value of $((\gamma)) + \Xi$ without knowing what $((\gamma))$ and the argument of $\Xi$ represent.  The expression is incomplete without definitions for these symbols.\\Final Answer: The final answer is $\boxed{undefined}$

\end{tcolorbox}
\caption{An example of model response from Gemini\_1.5 on MiP-Formula dataset. The model quickly identify the missing premise and abstain to answer.}
\label{fig:example-1-1}
\end{figure*}

\begin{figure*}[h]
\begin{tcolorbox}[
  enhanced, 
  colframe=cyan!75!black, 
  colback=white, 
  coltitle=white, 
  colbacktitle=cyan!75!black, 
  width=\linewidth, 
  arc=2mm, 
  auto outer arc, 
  boxrule=0.5pt, 
  left=10pt, 
  right=10pt, 
  drop shadow={black!50!white},
  top=10pt, 
  bottom=10pt, 
  title=\textbf{Model Response for a MiP-GSM8k Question from GPT-4o}, 
  fonttitle=\bfseries, 
  title code={ ode[rounded corners, fill=blue!75!black, draw=none, text=white] at (frame.title) {\textbf{xxx}};}, 
  attach boxed title to top center={yshift=-2mm}, 
  boxed title style={sharp corners, size=small}, 
]
\small
\textbf{Question:} Kyle bought last year's best-selling book for \$19.50. What was the original price of the book?

\textbf{Model Response:} 

To determine the original price of the book, we generally need to know the discount or markdown percentage applied to calculate its selling price of \$19.50. If Kyle bought the book for \$19.50 and we know it was discounted, we can use the following formula to find out the original price:\\\[ \text{Selling Price} = \text{Original Price} - (\text{Original Price} \times \text{Discount Rate}) \]\\Without additional information about discounts, promotions, or markdown rates applied, it's not possible to precisely determine the original price solely from the amount Kyle paid.\\If you provide the discount percentage or rate, I can help calculate the original price for you!

\end{tcolorbox}
\caption{An example of model response from GPT-4o on MiP-GSM8k dataset. The model quickly identify the missing premise and ask the user for more information.}
\label{fig:example-2-1}
\end{figure*}

\begin{figure*}[h]
\begin{tcolorbox}[
  enhanced, 
  colframe=orange!75!black, 
  colback=white, 
  coltitle=white, 
  colbacktitle=orange!75!black, 
  width=\linewidth, 
  arc=2mm, 
  auto outer arc, 
  boxrule=0.5pt, 
  left=10pt, 
  right=10pt, 
  drop shadow={black!50!white},
  top=10pt, 
  bottom=10pt, 
  title=\textbf{Model Response for a MiP-Formula Question from s1.1-32B}, 
  fonttitle=\bfseries, 
  title code={ ode[rounded corners, fill=blue!75!black, draw=none, text=white] at (frame.title) {\textbf{xxx}};}, 
  attach boxed title to top center={yshift=-2mm}, 
  boxed title style={sharp corners, size=small}, 
]
\small
\textbf{Question:} What is the value of $((\gamma)) + \Xi$?

\textbf{Model Response:}

Okay, so I need to figure out the value of $((\gamma)) + \Xi$. Hmm, first off, I should probably clarify what these symbols represent. The problem uses double parentheses around gamma, so (($\gamma$)), and then the Greek letter Xi ($\Xi$). I wonder if these are standard mathematical symbols or if they're defined in a specific context that I might not be aware of. \\Let me start by recalling what I know about gamma. In mathematics, $\gamma$ often refers to the Euler-Mascheroni constant, which is approximately 0.5772. It's a important constant in number theory and analysis, related to the harmonic series and the gamma function. But here, the notation is (($\gamma$)). The double parentheses could mean something specific. Maybe it's a notation from a particular field or textbook? For example, in some contexts, double parentheses might denote an interval or a floor function, but usually, floor is denoted by brackets like ($\gamma$). Alternatively, in physics, double parentheses could indicate some kind of tensor notation, but I'm not sure. \\Then there's the Xi ($\Xi$). Xi is another Greek letter, but in mathematics, $\Xi$ could stand for various things. In set theory, sometimes Xi is used as a variable for ordinals, or in other areas like particle physics, $\Xi$ represents certain particles. But again, without context, it's hard to say. Maybe it's a function or a constant defined in the problem's context? Wait, the question is just asking for the value of (($\gamma$)) + $\Xi$ without any additional information. That makes me think these symbols might be part of a puzzle or a play on notation rather than standard mathematical constants.\\Let me consider if this could be a trick question. The double parentheses around gamma(($\gamma$)) maybe that's referring to the number of letters in the word "gamma" or something like that? Let's see: Gamma is spelled G-A-M-M-A, which is 5 letters. So (($\gamma$)) could be 5. Then $\Xi$ is the Greek letter Xi, which is the 14th letter of the Greek alphabet. So maybe $\Xi$ represents 14? Then 5 + 14 would be 19. Is that possible? But is there a standard where Greek letters are assigned numerical values like Roman numerals? Wait, in Greek numerals, each letter does have a numerical value. Let me check that.\\In the Greek numeral system, letters are assigned values as follows:\\
\textbf{...............................................................67 more paragraphs here...............................................................}
\\But just to check another angle: In the context of APL programming language, certain symbols have specific meanings, but I don't think that's relevant here. \\Alternatively, in chemistry, $\gamma$ is used to denote surface tension, and $\Xi$ is not a standard symbol. In engineering, $\gamma$ is sometimes the shear strain. But again, not relevant. \\Alternatively, in statistics, gamma is a distribution, and Xi could be a parameter, but again, without context. \\Alternatively, the symbols might be from a specific equation in a paper or book, but since I don't have that reference, I can't confirm. \\Given all these considerations, and trying to think of what the answer might be expected to be in a problem like this, it's likely either 17 or 63. If the question is from a math competition or puzzle, they might be expecting the Greek numeral values. For example, in the Greek numeral system, $\gamma$ is 3 and $\Xi$ is 60, so 3 + 60 = 63. \\Alternatively, if it's based on their positions in the alphabet, 3 + 14 = 17. \\
\textbf{...............................................................21 more paragraphs here...............................................................}

Therefore, the answer is 63. The double parentheses around gamma might be a typo or just for decoration, but given the problem's phrasing, interpreting them as Greek numerals seems the most logical.\\**Final Answer**$\boxed{63}$ $</$think$>$To determine the value of (($\gamma$) + $\Xi$), we need to consider the possible meanings of the symbols ($\gamma$) and ($\Xi$). \\1. **Greek Numerals**: In the Greek numeral system, each letter has a specific numerical value. The letter ($\gamma$) (gamma) represents the value 3, and the letter ($\Xi$) (Xi) represents the value 60. 2. **Double Parentheses**: The notation (($\gamma$)) might be a red herring or simply emphasize that ($\gamma$) is being used as a numeral. The parentheses do not change the numerical value of ($\gamma$).\\Combining these values, we add the numerical values of ($\gamma$) and ($\Xi$): - ($\gamma$ = 3) - ($\Xi$ = 60)\\Thus, the value of (($\gamma$) + $\Xi$) is (3 + 60 = 63).\\$\boxed{63}$

\end{tcolorbox}
\caption{An example of response from s1.1 model on MiP-Formula data. The model spend lots of time doing inefficient and redundant reasoning before outputting a meaningless result.}
\label{fig:example-1-2}
\end{figure*}

\begin{figure*}[h]
\begin{tcolorbox}[
  enhanced, 
  colframe=orange!75!black, 
  colback=white, 
  coltitle=white, 
  colbacktitle=orange!75!black, 
  width=\linewidth, 
  arc=2mm, 
  auto outer arc, 
  boxrule=0.5pt, 
  left=10pt, 
  right=10pt, 
  drop shadow={black!50!white},
  top=10pt, 
  bottom=10pt, 
  title=\textbf{Model Response for a MiP-GSM8k Question from DeepSeek-R1}, 
  fonttitle=\bfseries, 
  title code={ ode[rounded corners, fill=blue!75!black, draw=none, text=white] at (frame.title) {\textbf{xxx}};}, 
  attach boxed title to top center={yshift=-2mm}, 
  boxed title style={sharp corners, size=small}, 
]
\small
\textbf{Question:} Kyle bought last year's best-selling book for \$19.50. What was the original price of the book?

\textbf{Model Response:} 

Okay, let's see. Kyle bought a book for \$19.50, and the question is asking for the original price. Hmm, wait a second. The problem mentions that it's last year's best-selling book, so maybe it was on sale or discounted? The question doesn't explicitly state that there was a discount, but it does say "last year's," which might imply that the book is now being sold at a reduced price. \\But hold on, the problem doesn't give any specific information about a discount percentage or a markdown. Without that, how can we calculate the original price? Maybe there's some context or common knowledge I'm missing here. For example, sometimes best-selling books are sold at a standard discount, but that's not a universal rule. \\Alternatively, maybe the question is testing understanding of something else, like sales tax or profit margins. But if it were sales tax, the original price would be less than \$19.50, and the total after tax would be more. But the question is asking for the original price, implying that \$19.50 is the price after some change. \\Wait, perhaps the problem is incomplete? The user just says "Kyle bought last year's best-selling book for \$19.50. What was the original price of the book?" If there's no additional information given, like a percentage discount or a mention that the price was reduced, then it's impossible to determine the original price. \\But maybe I need to make an assumption here. Since it's last year's best-selling book, it's common for such books to be discounted when a new edition comes out or after some time. Let me think, maybe the standard discount is 25\% or something? But the problem doesn't specify. \\Wait, perhaps the problem was part of a larger set where the discount rate was mentioned earlier, and this is a follow-up question. But as it stands, with the information given here, there's no way to calculate the original price. \\Alternatively, maybe the question is a trick question, where the original price is the same as what Kyle paid. But that doesn't make sense because it specifies "last year's best-selling book," which might imply it was sold at a different price before. \\Let me check the problem statement again: "Kyle bought last year's best-selling book for \$19.50. What was the original price of the book?" There's no mention of a discount, a sale, or any other financial context. \\If I were to approach this mathematically, let's denote the original price as P. If there was a discount, then \$19.50 is equal to P minus some discount. But without knowing the discount rate, we can't set up an equation. For example, if the book was discounted by 20\%, then the original price would be \$19.50 / 0.80. But since we don't have that percentage, we can't proceed.\\
\textbf{...............................................................24 more paragraphs here...............................................................}
\\Alternatively, if the answer is expected to be \$19.50, but that doesn't make sense because the question is asking for the original price, implying it was different. \\Alternatively, maybe the problem is in another language where "original price" refers to the price without tax, but in the US, sales tax isn't included in the price, so if \$19.50 is including tax, the original pre-tax price would be less. But again, without tax rate, can't compute. \\Alternatively, if the problem is from a country where tax is included, but the question is asking for the pre-tax price, but again, no tax rate given. \\This is really confusing. Since the problem is presented in English and mentions dollars, likely US context. In the US, sales tax is added at checkout, so the price Kyle paid would be the shelf price plus tax. If the question is asking for the pre-tax price, but again, no tax rate given. \\Alternatively, maybe the book was on sale, and the original price is the same as the sale price, but that contradicts the wording. \\Given all this, I think the most plausible answer, even though it's based on an assumption, is \$26, assuming a 25\% discount. Therefore, the original price was \$26.$<$think$>$ To determine the original price of the book, we need to consider that Kyle likely purchased it at a discounted price since it's "last year's best-selling book." A common assumption for such scenarios is a **25\% discount**. \\Let \( P \) be the original price. If the discounted price is \\\$19.50, this represents 75\% of the original price (100\% - 25\% discount):\\$[0.75P = 19.50]$\\Solving for \( P \):\\
$[P = \frac{19.50}{0.75} = 26]$
\\**Original price of the book:** \\$\boxed{26}$

\end{tcolorbox}
\caption{An example of model response from DeepSeek-R1 on MiP-GSM8k dataset. After thinking for a long time, the model hallucinates an answer based on its assumption of discount rate.}
\label{fig:example-2-2}
\end{figure*}
\clearpage

\end{document}